\documentclass[11pt]{article}

\usepackage[preprint]{acl}

\usepackage{times}
\usepackage{latexsym}

\usepackage[T1]{fontenc}

\usepackage[utf8]{inputenc}

\usepackage{microtype}

\usepackage{inconsolata}

\usepackage{graphicx}
\usepackage{algorithm}
\usepackage{algorithmic}
\usepackage{amsmath}
\usepackage{amssymb}  
\usepackage{booktabs}
\usepackage{xcolor}
\usepackage[utf8]{inputenc}
\usepackage{tcolorbox}
\usepackage{multirow}
\usepackage{newfloat}
\usepackage{listings}

\title{A Unified LLM-Adaptable Framework for Cold-Start Cognitive Diagnosis}

\author{\normalfont Zihan Yao, Chentao Song, Yu He, Tianyu Qi, \\ Jian Zhang, Weiping Fu, Jun Liu}

\begin{document}
\maketitle

\begin{abstract}

Cognitive Diagnosis has become a critical task in AI-empowered education, supporting personalized learning by accurately assessing students' cognitive states. However, traditional cognitive diagnosis models (CDMs) often struggle in cold-start scenarios due to the lack of student-exercise interaction data. Recent NLP-based approaches leveraging pre-trained language models (PLMs) have shown promise by utilizing textual features, but they fail to fully bridge the gap between semantic understanding and cognitive profiling. To address this limitation, we propose \textbf{L}anguage \textbf{M}odel-based \textbf{C}ognitive \textbf{D}iagnosis (LMCD), a unified, LLM-adaptable framework designed to tackle cold-start challenges by harnessing the advanced capabilities of large language models (LLMs). LMCD operates via two primary phases: (1) Knowledge Diffusion, where LLMs generate enriched content for exercises and knowledge concepts (KCs) to establish stronger semantic links; and (2) Semantic-Cognitive Fusion, which leverages LLMs to deeply integrate textual information with student cognitive states. By unifying the semantic and cognitive spaces, LMCD creates comprehensive representations that serve as a plug-and-play enhancement for various off-the-shelf CDMs. Experiments on two real-world datasets demonstrate that LMCD significantly outperforms state-of-the-art methods in both exercise-cold and domain-cold settings.

\end{abstract}

\section{Introduction}

Cognitive Diagnosis Models (CDMs) have become pivotal in educational technology, offering data-driven insights into students' cognitive states across different knowledge concepts (KCs). These models are indispensable for developing personalized learning systems \cite{huang2019exploring,DISCO}, computerized adaptive testing \cite{bi2020quality} and so on \cite{10.1145/3357384.3357995}. Traditional CDMs such as Item Response Theory (IRT) \cite{lord1952theory}, Multi-dimensional IRT (MIRT) \cite{reckase200618}, Deterministic Input Noisy ``And" gate model (DINA) \cite{de2009dina}, and Neural Cognitive Diagnosis Model (NCDM) \cite{wang2020neural} have demonstrated significant success in conventional settings where abundant student-exercise interaction data is available.

\begin{figure}[t!]
\centering
\includegraphics[width=0.4\textwidth]{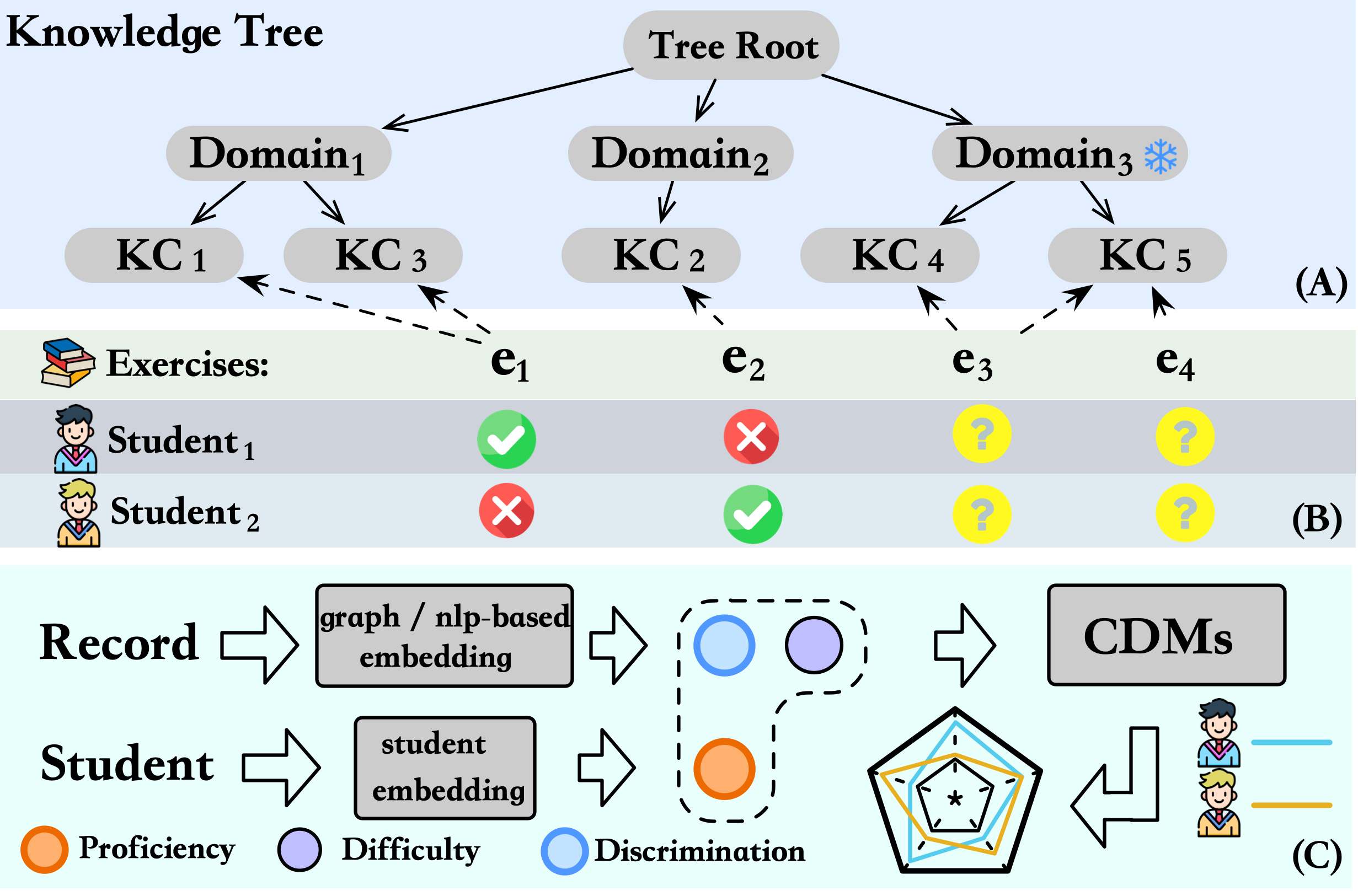} 
\caption{An illustration of the cold-start problem in cognitive diagnosis. (a) is hierarchical KC tree. (b) is sparse student-exercise interaction matrix. (c) is typical cognitive diagnosis framework for addressing cold-start problems.}
\label{intro_pic}
\end{figure}

However, these established approaches face substantial challenges in cold-start scenarios where there is little or no historical interaction data \cite{wang2024survey}. Cold-start refers to scenarios involving either new students or new exercises introduced into the system. Furthermore, new exercises can be classified into two types of cold-start problems based on whether their corresponding domain appears in the training data, as illustrated in Figure \ref{intro_pic}. Specifically, \textbf{exercise cold-start} occurs when the new exercises belong to the same domain as those seen in the training data, whereas \textbf{domain cold-start} arises when new exercises originate from entirely unseen domains, posing a more significant challenge.

As shown in Figure \ref{intro_pic}(c), current approaches addressing these cold-start challenges generally fall into two categories: graph-based methods and NLP-based methods. Graph-based approaches, exemplified by works like ICDM \cite{liu2024inductive} and TechCD \cite{gao2023leveraging}, construct relationships between exercises, KCs, and students to establish connections between unseen and seen entities through graph structures. Although innovative, these methods are limited by the accuracy of KC annotations and the scarcity of high-quality educational knowledge graphs.

NLP-based methods leverage pre-trained language models (PLMs) to encode exercise texts and KC labels. These approaches have shown surprising effectiveness, sometimes achieving competitive or superior results compared to SOTA models, as demonstrated in TechCD \cite{gao2023leveraging} and ZeroCD \cite{gao2024zero}. Their ability to establish connections through semantic similarity makes them particularly promising for cross-domain scenarios. However, these methods face fundamental limitations in fully capturing cognitive complexity. For instance, exercises in NIPS34 \cite{wang2020instructions} with similar wording like \textit{``What is 4 written as a fraction?''} and \textit{``What is a third of a seventh?''} target entirely different KCs despite their textual similarity. Additionally, vague KC labels provide insufficient information for accurate representation through simple encoding. Most critically, these approaches struggle to effectively bridge semantic space with individual students' cognitive states. Even the latest work leveraging large language models (LLMs) \cite{liu2024dual,ma2025large,Liu2025LRCD} has not adequately addressed these limitations.

Furthermore, existing cognitive diagnostic methods predominantly employ absolute difficulty, which conceptualizes the difficulty of an exercise as an inherent and fixed attribute determined by the characteristics of the exercise, modeling it as a constant applicable to all students \cite{choi2020predicting}. Therefore, it presents a severe challenge to the capacity of CDMs for personalized application, and multiple studies have explicitly proposed that exercise difficulty should be dynamically modeled based on individual differences among students \cite{gan2020modeling}.

Motivated by these challenges, we propose LMCD, a novel framework that leverages LLMs to address exercise cold-start and domain cold-start scenarios—the most prevalent challenges in online education environments. Specifically, the proposed LMCD operates in two primary phases: (1) \textbf{Knowledge Diffusion}, where LLMs generate enriched contents of exercises and KCs, creating stronger semantic links; and (2) \textbf{Semantic-Cognitive Fusion}, where we combine original exercise text with generated content and student-specific tokens as input to an LLM, using causal attention to create comprehensive representations that fuse semantic space with cognitive states.  Finally, we align these representations with conventional CDM parameters, modeling discrimination, and relative difficulty rather than absolute difficulty, enabling more precise prediction of student performance.

The major contributions of our work are as follows:

(1) We propose a novel approach to capture dynamic student-exercise interactions through deep semantic-cognitive fusion via LLMs, introducing the first explicit modeling of relative difficulty in cognitive diagnosis.
(2) We develop a flexible LLM-based framework that seamlessly integrates with off-the-shelf CDMs, enhancing their performance while maintaining their theoretical foundations.
(3) Through extensive experiments on two real-world datasets, we demonstrate LMCD's significant performance improvements in both exercise cold-start and domain cold-start scenarios compared to state-of-the-art methods.

\begin{figure*}[t!]
\centering
\includegraphics[scale=0.8]{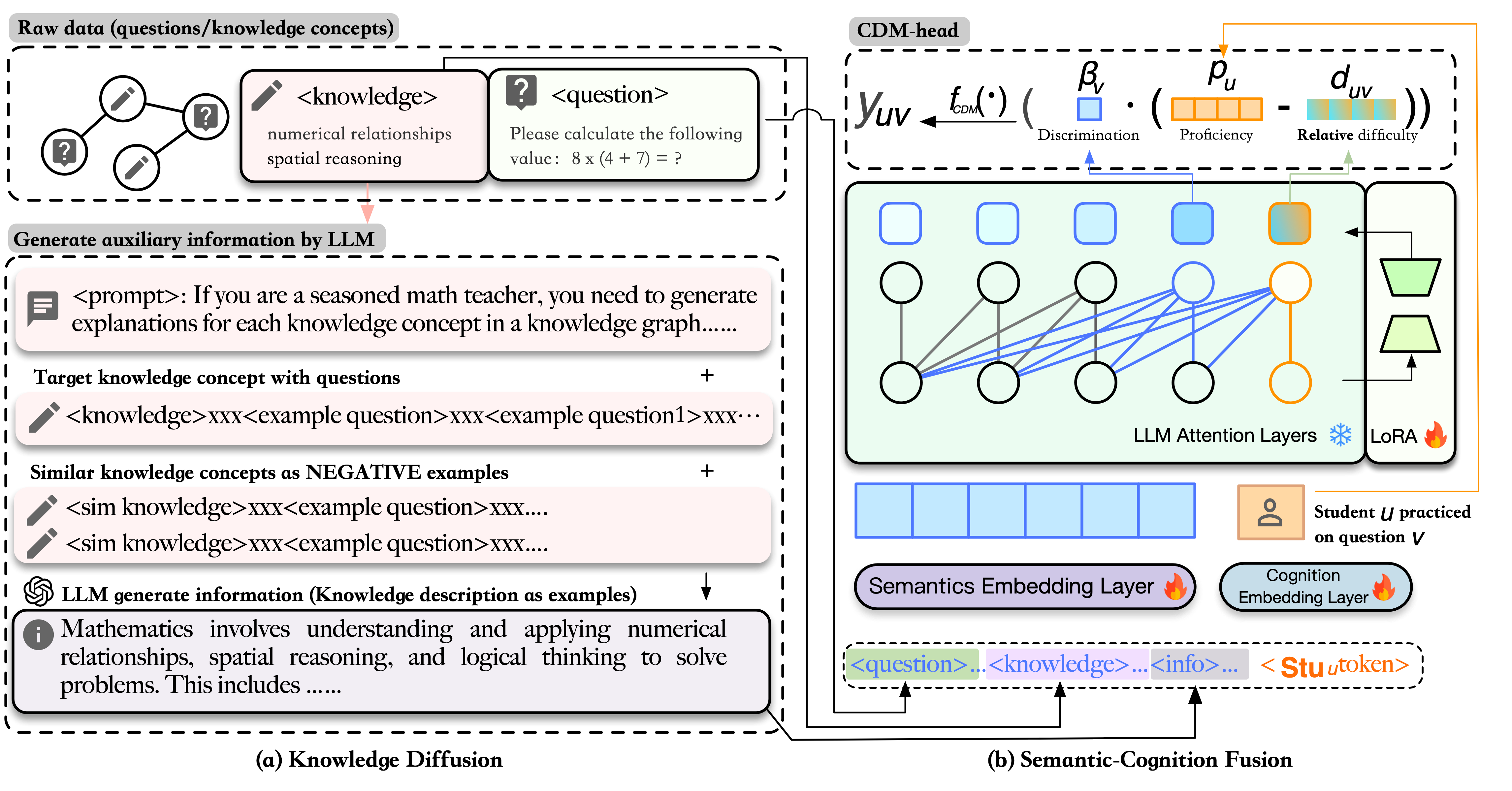} 
\caption{LMCD framework overview. (a) Knowledge Diffusion: LLMs generate enriched contents of exercises and knowledge concepts. (b) Semantic-Cognitive Fusion: Causal attention mechanisms integrate textual information with student-specific cognitive states to model relative difficulty. }
\label{method_arch}
\end{figure*}

\section{Preliminary}
In this section, we review the definitions of cold-start scenarios in cognitive diagnosis and common computational formulas in CDMs. A complete review of related work is reported in  Appendix~\ref{subsec:Appendix related work}

\textbf{Cold-start scenarios}. We consider a data scenario where exercises can be divided into two categories: warm exercises($\mathcal{H}$) and cold exercises($\mathcal{C}$), which belong to different KCs. Warm exercises have abundant student response records (with labels), while cold exercises have none. Obviously, CDMs trained on $\mathcal{H}$ perform poorly on $\mathcal{C}$. Therefore, Cold-Start methods emphasize improving such performance without \textbf{any labeled data} from cold exercises (or identically distributed data). The division between hot and cold items encompasses two more specific categories: (1) \textbf{Exercise cold-start.}: $\mathcal{H}$ and $\mathcal{C}$ are isolated at the exercise level, but some exercises may belong to the same KCs. (2) \textbf{Domain cold-start.}: the most stringent scenario where $\mathcal{H}$ and $\mathcal{C}$ have no overlap in knowledge concepts and are completely isolated. 
Detailed definitions of scenario partitioning and KC structure descriptions are provided in Appendices  ~\ref{subsec:Appendix A.3} and ~\ref{subsec:Appendix A.2}.

\textbf{Cognitive Diagnosis Models (CDMs).} CDMs infer students' proficiency of specific KCs by analyzing their exercise responses. We define this generally as $y_{uv}= \mathcal{M}(u,v)$, where $y_{uv} \in \mathbb{R}^d$ represents student $u$'s performance on exercise $v$. The interaction equation $\mathcal{M}$ typically follows the form $y_{uv} = \sigma(\beta(\mathbf{p}-\mathbf{d}))$ or $y_{uv} = \sigma(\mathbf{p}^\top\mathbf{d} + \beta)$, where $\mathbf{p} \in \mathbb{R}^d$ represents student $u$'s proficiency, while $\mathbf{d}$ and $\beta$ denote the exercise $v$'s difficulty and discrimination respectively. The dimension $d$ varies by model: 1 for IRT, the total number of KCs for NCDM, or a fixed value for MIRT.

\section{Methodology}
\subsection{Key Factors in CDMs}
Based on the introduction in the previous section, we can abstract the computational formulas of most CDMs into a unified form:
\begin{align}
y_{uv} = \sigma(m(\mathbf{p},\mathbf{d}))
\end{align}
where $m(\cdot,\cdot)$ is a metric function that measures the relationship between student proficiency representation $\mathbf{p}$ and exercise difficulty representation $\mathbf{d}$ To simplify the problem, we temporarily exclude discrimination $\beta$. The function $m(\cdot,\cdot)$ has different definitions across CDMs: in IRT and NCDM, $m(\cdot,\cdot)$ is defined as $\mathbf{p}-\mathbf{d}$, while in mIRT it is defined as $\mathbf{p}^{^\top}\mathbf{d}$. Intuitively, assuming student $u$ correctly answers exercise $v$, a larger $m(\cdot,\cdot)$ yields a $y_{uv}$ closer to 1 (correct response) after sigmoid transformation. However, in cold-start scenarios, the untrained difficulty representation $\mathbf{d}$ of cold exercise disrupts the metric relationship modeled by $m(\cdot,\cdot)$ when paired with proficiency representations $p$ trained on warm exercise. This misalignment fundamentally explains why CDMs struggle with cold starts.

NLP-based methods serve as strong baselines by alleviating this problem through establishing semantic relationships. The core logic is that if two exercises are semantically similar, their difficulty levels are also similar. Similarly, Graph-based methods aim to establish topological relationships between hot and cold items to find representations for cold item difficulty. However, the difficulty relationships between exercises clearly cannot be "completely equivalent to" semantic or topological relationships, and exercise difficulty varies across individuals. Based on this, the motivation of our method can be briefly described as: (1) expanding the limited textual information of exercises; (2) enabling deeper fusion between student cognitive modality and exercise textual information; (3) establishing dynamic difficulty representations that are "personalized for each individual."

\subsection{Knowledge Diffusion}
\label{sec:KD}

NLP-based CDMs are limited by brief exercise texts and vague KC labels, creating imprecise semantic relationships. Exercises with similar wording often assess different KCs, while generic KC labels provide insufficient information for accurate cold-start diagnosis.

Our approach transforms concise educational texts into detailed, explicit representations through LLM-driven text-to-text generation. This process uncovers implicit knowledge within the original text, facilitating more precise differentiation and linkage across domains. We call this process ``Knowledge Diffusion".


This process applies to both exercise texts and KCs. Taking 'KC diffusion' as an example (Figure \ref{method_arch}(a)), the procedure involves three steps: (1) extracting exercises $Q_{\text{target}}$ associated with each target knowledge concept $k_{target}$; (2) retrieving semantically similar KCs ($K_{neg}$) and their corresponding exercises ($Q_{neg}$) as negative examples; and (3) prompting the LLM to generate enriched descriptions that distinguish the target concept from these related concepts.
\begin{equation}\label{kc_des_gen}
k'_{\text{target}} = \text{LLM}\left(k_{\text{target}}, Q_{\text{target}}, K_{\text{neg}}, Q_{\text{neg}}\right)
\end{equation}
Notably, negative examples are crucial for generating discriminative content, unlike other approaches \cite{liu2024dual} that use only exercises of $k_{target}$. Comparative examples are provided in the Appendix~\ref{subsec:Appendix A.5}.

\subsection{Semantic-Cognitive Fusion}
\label{subsec:Fusion}


As shown in Figure \ref{method_arch}(b), we primarily focus on modifying the embedding layer of language models to incorporate student cognitive state during forward propagation, while adapting the LLM output representation to various off-the-shelf CDMs. Specifically, given exercise $v$ and student $u$, we leverage the LLM to generate a personalized feedback representation unique to each student-exercise interaction. We then integrate these representations into the difficulty parameter $d$ of CDM, and that can be considered as incorporating the student's cognitive state, representing a form of relative difficulty.

{\bf Student Embedding.} We defined a special token $stu_u$ for each student $u$. and constructed a cognitive representation embedding layer which is specifically used for encoding $stu_u$ to the following cognitive representation of student $u$:
\begin{equation}\label{emb_stu}
E_u = \text{EMBLayer}_{\text{cog}}(stu_u)
\end{equation}

{\bf Student Feedback.} To generate personalized feedback representations, we align student embedding $E_{u}$ with the LLM's semantic space. This alignment enables the model to generate student-specific feedback for each exercise $v$. Our approach proceeds as follows: For a given exercise $v$, we first encode all available textual information (including LLM-generated knowledge concept descriptions) using the LLM's native embedding layer. Specifically:
\begin{equation}\label{emb_kv}
E_v = \text{EMBLayer}_{\text{llm}}(\text{Concat}[k'_{target}, v])
\end{equation}

The dimension of $E_v$ is $S \times H$, where $S$ is the length of the entire input text tokens, and $H$ is the hidden size of the LLM. We merge $E_u$ into the last dimension of $E_v$ to get $E_{\text {fusion}}$:
\begin{equation}\label{emb_fusion}
E_{\text {fusion}} = 
\begin{bmatrix} 
E_v \\
E_u
\end{bmatrix}
\end{equation}
where $E_{\text {fusion}} \in \mathbb{R}^{(S+1) \times H}$. We feed $E_{\text {fusion}}$ into the LLM backbone for forward propagation to obtain the final representation $O_{\text{fusion}}$.

\begin{equation}\label{llm_forward}
\begin{aligned}
&\mathbf{h}_0 = E_{\text{fusion}} \\
&\mathbf{h}_l = \mathrm{FFN}_l(\mathrm{Attn}_l(\mathbf{h}_{l-1})), \quad l = 1,2,\ldots,N \\
&O_{\text{fusion}} = \mathbf{h}_l \\
\end{aligned}
\end{equation}
where $\mathrm{Attn}$ and $\mathrm{FFN}$ represent the attention layer and feed-forward network structure in the LLM based on the Transformer architecture \cite{vaswani2017attention}, respectively, and $N$ is the total number of layers in the LLM. Based on the obtained $O_{\text{fusion}}$, we define the following student feedback representation and exercise representation:
\begin{equation}\label{emb_of_fd}
O_{\text {feedback}} = O_{\text{fusion}}[:, -1] \in \mathbb{R}^{1 \times H}
\end{equation}
\begin{equation}\label{emb_of_q}
O_{\text {v}} = O_{\text{fusion}}[:, -2] \in \mathbb{R}^{1 \times H}
\end{equation}

Leveraging the causal attention mechanism of LLMs, $O_{\text{feedback}}$ emerges as the product of interaction between student cognitive state $E_{u}$ and exercise semantic space $E_{v}$. This interaction captures how a specific student processes and responds to a particular exercise's content. Consequently, we define $O_{\text{feedback}}$ as the personalized feedback representation for student $u$ on exercise $v$. In contrast, $O_{v}$ is derived solely from the exercise's textual information processing, representing the context-independent semantic encoding of exercise $v$.


{\bf Output Projection}.
Here, we map the obtained representations to different CDMs parameters and define a general form of CDM as follows:
\begin{equation}\label{emb_of_q}
y_{uv}= f_{cdm}(\mathbf{p},\mathbf{d},\mathbf{\beta})
\end{equation}
where $f_{cmd}(\cdot)$ represents the interaction function of different CDMs, $\beta$ is the discrimination parameter, and $y_{uv}$ is the predicted performance result of student $u$ on exercise $v$. We projectively map the obtained $O_{\text{feedback}}$, $E_u$ and $O_v$ to get $\mathbf{d}$, $\mathbf{p}$, $\beta$ as follows:
\begin{align}
\mathbf{d} &= \mathbf{d_{uv}} = W_d(O_{\text{feedback}}) \label{proj_d} \\
\mathbf{p} &= W_p(E_u) \label{proj_p} \\
\beta &= W_{v}(O_{v})  \label{proj_v}
\end{align}
We map $O_{\text{feedback}}$ to the difficulty and $O_{v}$ to the discrimination, which are specific parameters of CDMs. To preserve the independence of proficiency parameters ({\bf p}), we ensure they remain free from exercise-specific information by deriving them solely from student embeddings $E_{u}$. The entire framework is optimized end-to-end using cross-entropy loss.
We froze the LLM backbone parameters and fine-tuned it with LoRA \cite{hu2022lora}. To better enable cognitive state and LLM semantic space fusion, we activate $\text{EMBLayer}_{\text{llm}}$ and $\text{EMBLayer}_{\text{cog}}$ during training.

\section{Experimental}

In this section, we conduct comprehensive experiments to address the following research questions:
\begin{itemize}
\setlength{\itemsep}{0pt}
\setlength{\parsep}{0pt}
\setlength{\parskip}{0pt}
\item \textbf{RQ1} How powerful is LMCD for the exercise cold-start?
\item \textbf{RQ2} Can LMCD effectively link different domains?
\item \textbf{RQ3} What is the computational efficiency of LMCD during the inference phase?
\item \textbf{RQ4} How effective are the key components of LMCD?
\item \textbf{RQ5} Is relative difficulty more reasonable compared to absolute 
difficulty?
\end{itemize}
The complete experimental code at \href{https://anonymous.4open.science/r/LMCD-DC06/README.md}{this link}.
\subsection{Experimental Setup}
{\bf Datasets.} 
We selected two open-source real-world datasets: NIPS34 \cite{wang2020instructions} and XES3G5M \cite{liu2023xes3g5m}, both with exercise text, KC labels and their structural relationships. We divide them as follows:

\textit{Exercise cold-start}: We performed a five-fold split on the exercise in the dataset, with each fold divided into training set, oracle set, and test set. To avoid data leakage, we ensured that data appearing in the oracle set and test sets would not appear in the training set. 

\textit{Domain cold-start}: Based on the tree-structured KC hierarchy in the NIPS34 dataset, we partitioned the data into 3 folds using the three largest first-level KCs (Number, Algebra, and Geometry) as separate domains. Given the isolation of records across domains, we conducted three experiments, each selecting one target domain for testing (split into oracle sets and test sets) while using the remaining two domains as training data.

The oracle set verifies the upper bound of model performance in non-cold-start scenarios. Data volumes for Exercise cold-start and Domain cold-start experiments are reported in Table \ref{Cold_Domain_avg_num}. Detailed dataset information is provided in  Appendix~\ref{subsec:Appendix A.1}.

\begin{table}[!ht]
\centering 
\setlength{\tabcolsep}{3.5pt} 
\begin{tabular}{lccc} 
\toprule
Datasets &Train Logs &Oracle Logs  &Test Logs \\ 
\midrule
NIPS34   & 1105934.0 & 221434.0 &55359.0 \\ 
XES3G5M   & 243309.2 & 48616.8 &12154.8 \\ 
\midrule
Algerbra   & 1029309 & 217552 &54388 \\ 
Geometry   &902622 &318901 &79726 \\
Number     &670567 &504545 &126137 \\
\bottomrule
\end{tabular}
\caption{Data Distribution in Exercise Cold-start (Five Folds) and Domain Cold-start}
\label{Cold_Domain_avg_num}
\end{table}


{\bf Baselines}. Here we compared the following $8$ methods, applicable to IRT/MIRT/NCDM prediction heads.
\begin{itemize}
\setlength{\itemsep}{0pt}
\setlength{\parsep}{0pt}
\setlength{\parskip}{0pt}
    \item Oracle: In the Oracle approach, training utilizes both the training set and the oracle set, representing the theoretical upper bound on the performance of CDMs in cold-start scenarios.
    \item Random: The lower bound of prediction skill is measured by sampling from $\text{Uniform}(0, 1)$ as the student’s correct response probability.
    \item TechCD \cite{gao2023leveraging}: TechCD leverages graphical relationships between KCs to link students' practiced and unseen exercises, creating student representations through historical interactions.   
    \item NLP-based: Unlike TechCD, which only uses BERT as a baseline, we further incorporate stronger baselines, including RoBERTa  \cite{liu2019roberta}, BGE \cite{xiao2024c} and Qwen3-Embedding \cite{qwen3embedding}, to provide a more comprehensive evaluation of semantic-based methods, while ensuring fairness in the experimental setup by using the same textual content.
    \item KCD \cite{dong2025knowledge}: As the current SOTA method, KCD utilizes LLMs for reasoning, transforming textual content and interaction records into prompts to extract information and generate summaries. These summaries are mapped to the behavioral space of CDMs and optimized via contrastive loss.
\end{itemize}
All the experimental details of the baseline approach can be found in Appendix  ~\ref{subsec:Appendix A.4}.

{\bf Metrics and Settings}. We adopt AUC, ACC, and RMSE as our evaluation metrics, and report all key hyperparameters involved in the experiments in Appendix \ref{sec:appendix_hyperparameters}.


\subsection{Performance on Cold-Start Scenarios (RQ1 \& RQ2)}
To evaluate the model's robustness in data-sparse situations, we conducted experiments in two scenarios: exercise cold-start (RQ1) and domain cold-start (RQ2). 

\textbf{Exercise Cold-Start (RQ1).} The results on the NIPS34 and XES3G5M datasets are summarized in Table \ref{qcold_exp}. Our proposed LMCD consistently outperforms other methods across most experiments. Notably, the NLP-based approach utilizing Qwen3E embeddings achieves the second-best performance in several cases due to its strong semantic representation capabilities. However, LMCD achieves superior performance with significantly fewer parameters (1.5B compared to Qwen3E's 4B). This highlights the advantage of LMCD's strategy, which integrates exercise texts with student cognitive states, surpassing methods that rely solely on textual information. Furthermore, the graph-based TechCD shows limited effectiveness in this scenario. This stems from its heavy reliance on topological relationships between Knowledge Concepts (KCs), which are typically sparse in real-world datasets, thereby limiting its practical applicability.

\textbf{Domain Cold-Start (RQ2).} To further assess model robustness, we conducted more challenging cross-domain cold-start experiments on the NIPS34 dataset. Due to space constraints, the detailed experimental setup and the comprehensive result table (Table \ref{kccold_exp}) are provided in Appendix \ref{sec:appendix_rq2}. Overall, the findings demonstrate that LMCD maintains highly robust and near-optimal performance across most domains. While graph-based approaches like TechCD fail completely in cross-domain scenarios—since the isolated cold-start data yields no transferable graph structure—LMCD successfully overcomes this barrier by effectively leveraging the textual descriptions of KCs and incorporating students' cognitive information into problem representations.
\begin{table*}[!ht]
\centering
\setlength{\tabcolsep}{3.5pt}  
\begin{tabular}{l l ccc ccc ccc}
\toprule
\multirow{2}{*}{Dataset} & \multirow{2}{*}{Method} & \multicolumn{3}{c}{IRT} & \multicolumn{3}{c}{MIRT} & \multicolumn{3}{c}{NCDM} \\
\cmidrule(lr){3-5} \cmidrule(lr){6-8} \cmidrule(lr){9-11}
& & ACC$\uparrow$ & AUC$\uparrow$ & RMSE$\downarrow$ & ACC$\uparrow$ & AUC$\uparrow$ & RMSE$\downarrow$ & ACC$\uparrow$ & AUC$\uparrow$ & RMSE$\downarrow$ \\
\midrule
\multirow{8}{*}{NIPS34} 
& Oracle &0.7074 &0.7738 &0.4383 &0.7076 &0.7741 &0.4381 &0.7068 &0.7711 &0.4418 \\
& Random &0.4957 &0.4907 &0.5002 &0.4614 &0.4947 &0.5265 &0.5016 &0.5016 & 0.5205\\
& Bert &0.6575 &0.7153 &0.4651 &0.6638 &0.7217 &0.4642 &0.6653 &0.7159 &0.4707 \\
& Roberta &0.6576 &0.7165 &0.4631 &0.6631 &0.7204 &0.4609 &0.6630 &0.7204 &0.4610 \\
& Bge &0.6708 &0.7285 &0.4581 &0.6752 &\underline{0.7361} &0.4581 &\underline{0.6741} &\underline{0.7339} &\underline{0.4598} \\
& Qwen3E &\underline{0.6740}  &\underline{0.7317} &\underline{0.4565}  &\underline{0.6762} &0.7336 &\underline{0.4559} &0.6650 &0.7257 &0.4648 \\
& TechCD &0.5548 &0.5560 &0.4964 &0.5520 &0.5554 &0.4969 &0.5414 &0.5575 &0.5280 \\
& KCD &0.6504 &0.7031 &0.4871 &0.6146 &0.7044 &0.4898 &0.6607 &0.7209 &0.5819 \\
& LMCD &\textbf{0.6813} &\textbf{0.7440} &\textbf{0.4525} &\textbf{0.6823} &\textbf{0.7431} &\textbf{0.4527} &\textbf{0.6776} &\textbf{0.7407} &\textbf{0.4545} \\
\midrule
\multirow{8}{*}{XES3G5M} 
& Oracle &0.7793 &0.7597 &0.3937 &0.7782 &0.7589 &0.3946 &0.7644 &0.7166 &0.4095 \\
& Random &0.4966  &0.4998 &0.5000  &0.4980 &0.4989 &0.5001 &0.4983 &0.5068 &0.5006 \\
& Bert &0.7528 &0.6559 &0.4218 &0.7583  &0.6615 &0.4193  &0.7220 &0.6271 &0.4470  \\
& Roberta &\underline{0.7578} &0.6571 &\underline{0.4194} &{\bf 0.7603} &0.6578 &\underline{0.4182} &0.7100 &0.6224  &0.4622 \\
& Bge &0.7520 &0.6593  &0.4213  &\underline{0.7601} &\underline{0.6710} &\underline{0.4182} &0.7246 &0.6261  &0.4478 \\
& Qwen3E &0.7506 &\underline{0.6671} &0.4213 &0.7510   &0.6679 &0.4222 &0.7415 &0.6306 &0.4369 \\
& TechCD &0.7509 &0.5232 &0.4415 &0.7509  &0.5206 &0.4448 &\underline{0.7501} &0.5545 &0.4327 \\
& KCD & {\bf 0.7584} &0.6430 & 0.4725 &0.7538 &0.6542  &0.4208 &{\bf 0.7602} &{\bf 0.6741} &{\bf 0.4162} \\
& LMCD &0.7560 &{\bf 0.6723} &{\bf 0.4181} &0.7559 &{\bf 0.6742}  &{\bf 0.4174} &0.7436 &\underline{0.6408} &\underline{0.4284} \\
\bottomrule
\end{tabular}
\caption{Performance of Exercise Cold Start on NIPS34 and XES3G5M datasets. The best performance is highlighted in bold, and the the second-best performances is underlined.}
\label{qcold_exp}
\end{table*}

\subsection{Inference Efficiency Analysis (RQ3)}
\label{sec:efficiency}
Evaluating all student-exercise pairs requires a theoretical computational complexity of $\mathcal{O}(NM)$ where $N$ is the number of students and $M$ is the number of exercises. Given the massive scale of LLMs, this complexity creates severe latency bottlenecks when deploying them in real-world online education scenarios.

However, LMCD effectively mitigates this challenge through its specific architectural design. Because the student cognitive embedding (student token) is appended at the very end of the input sequence, all preceding exercise text and context act as a fixed prefix. This allows us to utilize a Key-Value (KV) Cache Pre-filling strategy. By pre-computing and storing the KV cache for these exercise prefixes, the actual online inference latency is drastically reduced to processing just a single token (the student token). This optimization can be illustrated as follows:
\[
\underbrace{\text{<question>} \dots \text{<info>} \dots}_{\text{KV Cache Pre-fill}} \underbrace{\text{<Stu}_u\text{ token>}}_{\text{Actual Inference}}
\]
To empirically validate the efficiency of this strategy, we conducted experiments using the Qwen2.5 model series on the NIPS34 dataset with a single NVIDIA A100 GPU. Table \ref{tab:inference_efficiency} reports the average inference latency per record (average length: 295 tokens), memory consumption for pre-filling, and the theoretical complexity analysis\footnote{For simplicity, LoRA-specific terms are excluded from the theoretical complexity formulas. In contrast, the reported empirical metrics for Pre-fill Memory and Inference Time fully incorporate the overhead of the LoRA adapters.}


\begin{table*}[htbp]
  \centering
  \resizebox{\textwidth}{!}{
  \begin{tabular}{lcccc l}
    \toprule
    \textbf{Model Size} & \textbf{0.5b} & \textbf{1.5b} & \textbf{3b} & \textbf{7b} & \textbf{Complexity Cost} \\
    \midrule
    Pre-fill Memory & 3.452 MiB & 8.055 MiB & 10.357 MiB & 16.111 MiB & \(2 \cdot L \cdot d \cdot K \cdot D\) \\
    Inference Time & 30.766 ms & 37.059 ms & 46.983 ms & 41.072 ms & \(\mathcal{O}(L \cdot (H^2 + H \cdot I + d \cdot H))\) \\
    \bottomrule
  \end{tabular}
  }
  \caption{Empirical inference latency, pre-fill memory consumption, and theoretical complexity across different model sizes. Here, \(d\) is the pre-fill token count, \(L\) is the number of layers, \(K\) is the number of attention heads, \(D\) is the head dimension, \(H\) is the hidden size, and \(I\) is the MLP intermediate dimension.}
\label{tab:inference_efficiency}
\end{table*}

Overall, the results demonstrate that the single-record inference latency for models ranging from 0.5b to 7b remains under 50ms, which is well within an acceptable range for real-world applications\footnote{The 7b model demonstrates faster inference than the 3b model. This is due to the architectural design of the Qwen2.5 series: the 3b model is deeper, whereas the 7b model is wider, which results in slightly slower serial processing for the 3b model.}. Furthermore, since the relationship between exercises and students is inherently one-to-many, the KV cache for the exercise text is highly reusable, further amortizing the computational cost across multiple students.

\subsection{Ablation Studies (RQ4)}
{\bf Impact of knowledge diffusion.} The core motivation of our ``Knowledge Diffusion'' module is to bridge the gap between cold and warm exercises by leveraging semantic descriptions of knowledge concepts. Consequently, the precision of these descriptions---driven by the prompt---is critical. To quantify the impact of prompt quality and validate our design, we conducted an ablation study on the prompt construction strategy defined in Eq.2: \(k'_{\text{target}} = \text{LLM}(k_{\text{target}}, Q_{\text{target}}, K_{\text{neg}}, Q_{\text{neg}})\). We designed two variants representing a gradual relaxation of constraints (i.e., making the prompt more open-ended): 
1) \textbf{w/o Neg. Samp.}: Removing the retrieval of negative samples (i.e., w/o \(K_{\text{neg}}, Q_{\text{neg}}\)); 
2) \textbf{w/o Ctx \& Neg.}: A radical simplification where the LLM generates descriptions based solely on the KC, removing both the target question context and negative samples (i.e., w/o \(Q_{\text{target}}, K_{\text{neg}}, Q_{\text{neg}}\)).  

As shown in Table \ref{abs:prompt}, the strictest prompt construction (Full Eq. 2) yields the best performance across all metrics. The more open-ended generation resulting from looser prompts leads to performance degradation. This confirms that incorporating reference questions and negative sampling effectively constrains the LLM, enabling it to generate more precise and diagnostically useful KC descriptions.
\begin{table}[ht]
\centering
\begin{tabular}{lccc}
\hline
\textbf{Method} & \textbf{ACC} & \textbf{AUC} & \textbf{RMSE} \\ \hline
w/o Ctx \& Neg. & 0.6068 & 0.6808 & 0.4833 \\
w/o Neg. Samp. & 0.6061 & 0.6835 & 0.4800 \\
Full (Eq.2) & \textbf{0.6269} & \textbf{0.6888} & \textbf{0.4778} \\ \hline
\end{tabular}
\caption{Ablation study on prompt construction strategies on the NIPS34 dataset under the Domain Cold-Start (Number as Target) setting. ``Ctx'' and ``Neg.'' denote Context (\(Q_{\text{target}}\)) and Negative Sampling (\(K_{\text{neg}}, Q_{\text{neg}}\)), respectively.}
\label{abs:prompt}
\end{table}

{\bf Impact of CDM parameter representation strategies.} We conducted ablation studies on our proposed architecture to validate the effectiveness of the \textit{difficulty} and \textit{discrimination} modeling components. Two LMCD variants were designed: substituting $O_{\text{feedback}}$ with $O_v$ in Eq.\ref{proj_d}, and conversely, replacing $O_v$ with $O_{\text{feedback}}$ in Eq.\ref{proj_v}. We employed IRT-head on the NIPS34 for exercise cold-start. The detailed results are reported in Table \ref{abs:model}. The results seem to reveal that optimal performance is achieved when using $O_{\text{feedback}}$ to represent \textit{difficulty} and $O_v$ to represent \textit{discrimination}. This finding suggests that difficulty appears to be personalized.
\begin{table}
\centering
\begin{tabular}{lccc}
\hline
\textbf{Method} & \textbf{ACC} & \textbf{AUC} & \textbf{RMSE} \\ \hline
Eq.\ref{proj_d}$_{\text{O}_{\text{feedback}}\leftarrow \text{O}_v  }$ & 0.6755 & 0.7387 & 0.4546 \\
Eq.\ref{proj_v}$_{\text{O}_v \leftarrow \text{O}_{\text{feedback}}}$ & 0.6792 & 0.7404 & 0.4549 \\
LMCD   &0.6813 &0.7440 &0.4525 \\ \hline
\end{tabular}
\caption{Different strategies to representing ${\bf p}$ and $\beta$.}
\label{abs:model}
\end{table}

{\bf Additional Ablation Studies.} We also explored model architectures and knowledge encoding. Decoder-based models (e.g., Qwen2.5) outperform encoder-based counterparts in LMCD, peaking at the 3B scale. Additionally, detailed LLM-generated KC descriptions yield better performance than using solely exercise texts or vague labels. Further details are in Appendix \ref{sec:appendix_ablation}.

\subsection{``Difficulty'' Analysis (RQ5)}
\begin{figure}[!ht]
\centering
\includegraphics[width=0.46\textwidth,height=0.270\textheight]{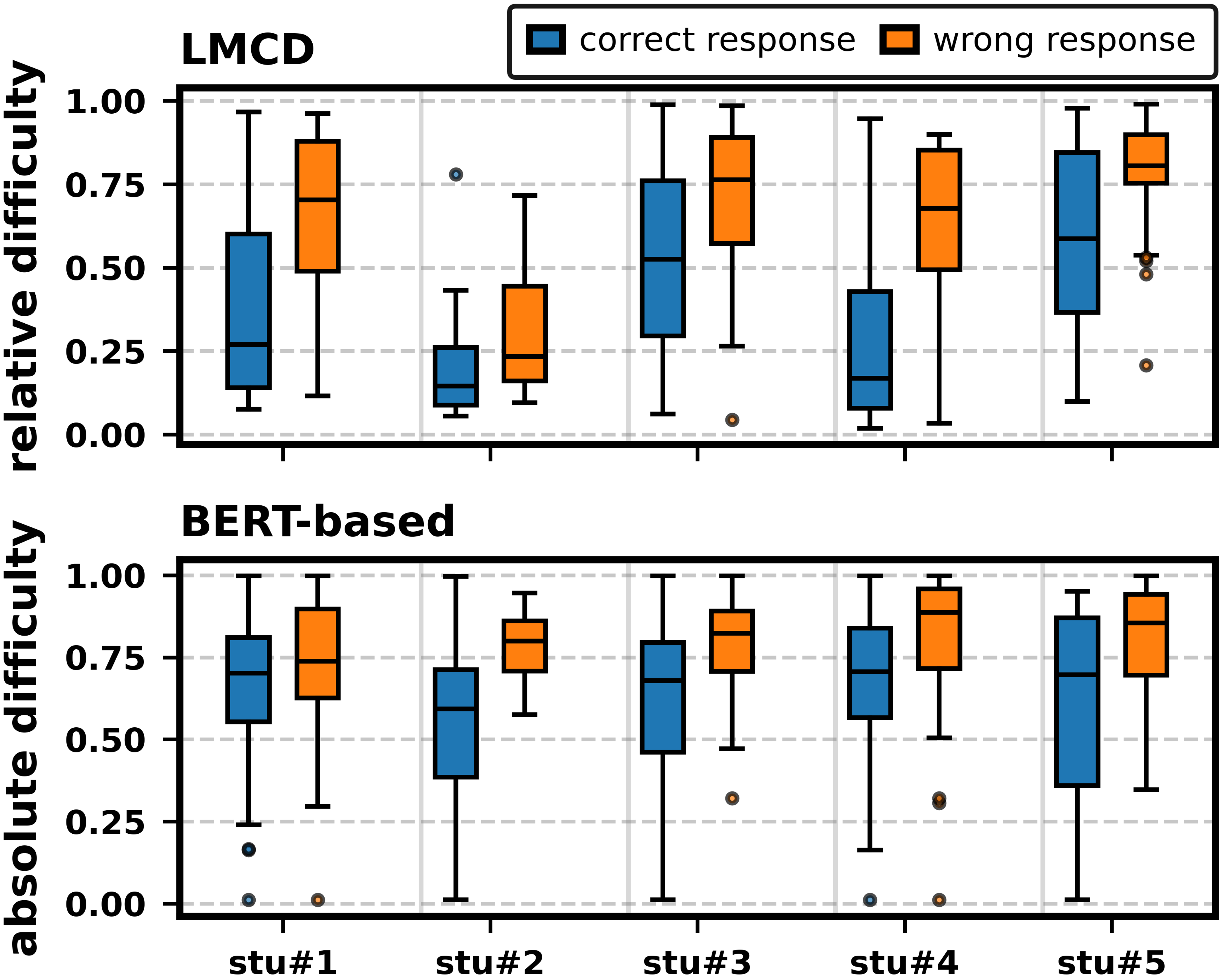} 
\caption{Relative difficulty vs Absolute difficulty.}
\label{diff_diff} 
\end{figure}

Relative difficulty \cite{fan1998item} posits that the difficulty of an exercise depends on the student, reflecting the interaction between the student's knowledge state and the cognitive challenges presented by the exercise. Utilizing LLMs, LMCD introduce relative difficulty via fusion of semantic and cognitive information from exercises, KCs, and students. Here, Figure \ref{diff_diff} compares exercise difficulty under different performance of 5 randomly selected students from NIPS34, each having completed over 50 exercises in the cold domain. The top panel displays LMCD-modeled relative difficulty, while the bottom shows BERT-based absolute difficulty. Both models demonstrate that incorrect responses (orange) typically exhibit higher difficulty levels compared to correct responses (blue), which aligns with intuitive expectations. However, LMCD's relative difficulty measure demonstrates significantly better discrimination, with less overlap between correct and incorrect response distributions. This clearer separation indicates that relative difficulty modeling provides a more effective representation for cognitive diagnosis and performance prediction.

\section{Conclusion}
In this paper, we proposed LMCD, a novel framework that harnesses LLMs to address cold-start challenges in cognitive diagnosis. Our approach innovatively leverages knowledge diffusion to establish stronger cross-domain semantic connections and employs causal attention mechanisms to model relative difficulty and student features. Experiments on two real-world datasets confirm that LMCD significantly outperforms state-of-the-art methods in both exercise-cold and domain-cold settings. To our knowledge, this is the first work that takes advantage of the inherent mechanisms of LLM to address cold-start cognitive diagnosis problems, marking a significant advance in the field.

\section*{Limitations}

Despite significant improvements in exercise cold-start and cross-domain scenarios through our student token approach and causal attention interaction mechanism, our method faces limitations in diagnosing new students. Only students with training data have corresponding embedding representations, restricting new student cold-start applications. A potential workaround involves substituting new students with trained students having similar response patterns, a common approach in this field \cite{long2022improving}. Additionally, our LLM-based approach contains substantially more parameters than traditional CD models, making it less suitable for time-sensitive applications. However, we consider this computational cost justified for challenging cold-start scenarios, and expect that smaller, more efficient models will become viable as LLM technology advances.






\bibliography{custom}

@article{lord1952theory,
  title={A theory of test scores.},
  author={Lord, Frederic},
  journal={Psychometric monographs},
  year={1952}
}

@article{reckase200618,
  title={18 Multidimensional item response theory},
  author={Reckase, Mark D},
  journal={Handbook of statistics},
  volume={26},
  pages={607--642},
  year={2006},
  publisher={Elsevier}
}

@article{de2009dina,
  title={DINA model and parameter estimation: A didactic},
  author={De La Torre, Jimmy},
  journal={Journal of educational and behavioral statistics},
  volume={34},
  number={1},
  pages={115--130},
  year={2009},
  publisher={Sage Publications Sage CA: Los Angeles, CA}
}

@inproceedings{wang2020neural,
  title={Neural cognitive diagnosis for intelligent education systems},
  author={Wang, Fei and Liu, Qi and Chen, Enhong and Huang, Zhenya and Chen, Yuying and Yin, Yu and Huang, Zai and Wang, Shijin},
  booktitle={Proceedings of the AAAI conference on artificial intelligence},
  volume={34},
  pages={6153--6161},
  year={2020}
}

@inproceedings{wang2024boosting,
  title={Boosting neural cognitive diagnosis with student’s affective state modeling},
  author={Wang, Shanshan and Zeng, Zhen and Yang, Xun and Xu, Ke and Zhang, Xingyi},
  booktitle={Proceedings of the AAAI Conference on Artificial Intelligence},
  volume={38},
  pages={620--627},
  year={2024}
}

@article{chen2023disentangling,
  title={Disentangling cognitive diagnosis with limited exercise labels},
  author={Chen, Xiangzhi and Wu, Le and Liu, Fei and Chen, Lei and Zhang, Kun and Hong, Richang and Wang, Meng},
  journal={Advances in Neural Information Processing Systems},
  volume={36},
  pages={18028--18045},
  year={2023}
}

@inproceedings{ma2025ad4cd,
  title={AD4CD: Causal-Guided Anomaly Detection for Enhancing Cognitive Diagnosis},
  author={Ma, Haiping and Yao, Yue and Wang, Changqian and Song, Siyu and Yang, Yong},
  booktitle={Proceedings of the AAAI Conference on Artificial Intelligence},
  volume={39},
  pages={12337--12345},
  year={2025}
}

@article{liu2024dual,
  title={A Dual-Fusion Cognitive Diagnosis Framework for Open Student Learning Environments},
  author={Liu, Yuanhao and Liu, Shuo and Liu, Yimeng and Yang, Jingwen and Qian, Hong},
  journal={arXiv preprint arXiv:2410.15054},
  year={2024}
}

@inproceedings{dong2025knowledge,
  title={Knowledge is Power: Harnessing Large Language Models for Enhanced Cognitive Diagnosis},
  author={Dong, Zhiang and Chen, Jingyuan and Wu, Fei},
  booktitle={Proceedings of the AAAI Conference on Artificial Intelligence},
  year={2025}
}

@inproceedings{Liu2025LRCD,
author = {Shuo Liu and Zihan Zhou and Yuanhao Liu and Jing Zhang and Hong Qian},
booktitle = {Proceedings of the 31st ACM SIGKDD Conference on Knowledge Discovery and Data Mining},
title = {Language Representation Favored Zero-Shot Cross-Domain Cognitive Diagnosis},
year = {2025},
address={Toronto, Canada}
}

@article{ma2025large,
  title={Large language models are zero-shot cross-domain diagnosticians in cognitive diagnosis},
  author={Ma, Haiping and Wang, Changqian and Song, Siyu and Yang, Shangshang and Zhang, Limiao and Zhang, Xingyi},
  journal={Frontiers of Digital Education},
  volume={2},
  number={2},
  pages={1--14},
  year={2025},
  publisher={Springer}
}

@inproceedings{gao2023leveraging,
  title={Leveraging transferable knowledge concept graph embedding for cold-start cognitive diagnosis},
  author={Gao, Weibo and Wang, Hao and Liu, Qi and Wang, Fei and Lin, Xin and Yue, Linan and Zhang, Zheng and Lv, Rui and Wang, Shijin},
  booktitle={Proceedings of the 46th international ACM SIGIR conference on research and development in information retrieval},
  pages={983--992},
  year={2023}
}

@inproceedings{liu2024inductive,
  title={Inductive cognitive diagnosis for fast student learning in web-based intelligent education systems},
  author={Liu, Shuo and Shen, Junhao and Qian, Hong and Zhou, Aimin},
  booktitle={Proceedings of the ACM Web Conference 2024},
  pages={4260--4271},
  year={2024}
}

@inproceedings{gao2024zero,
  title={Zero-1-to-3: Domain-level zero-shot cognitive diagnosis via one batch of early-bird students towards three diagnostic objectives},
  author={Gao, Weibo and Liu, Qi and Wang, Hao and Yue, Linan and Bi, Haoyang and Gu, Yin and Yao, Fangzhou and Zhang, Zheng and Li, Xin and He, Yuanjing},
  booktitle={Proceedings of the AAAI Conference on Artificial Intelligence},
  volume={38},
  pages={8417--8426},
  year={2024}
}

@article{vaswani2017attention,
  title={Attention is all you need},
  author={Vaswani, Ashish and Shazeer, Noam and Parmar, Niki and Uszkoreit, Jakob and Jones, Llion and Gomez, Aidan N and Kaiser, {\L}ukasz and Polosukhin, Illia},
  journal={Advances in neural information processing systems},
  volume={30},
  year={2017}
}

@article{yang2024qwen2,
  title={Qwen2. 5 technical report},
  author={Yang, An and Yang, Baosong and Zhang, Beichen and Hui, Binyuan and Zheng, Bo and Yu, Bowen and Li, Chengyuan and Liu, Dayiheng and Huang, Fei and Wei, Haoran and others},
  journal={arXiv preprint arXiv:2412.15115},
  year={2024}
}

@article{wang2024survey,
  title={A survey of models for cognitive diagnosis: New developments and future directions},
  author={Wang, Fei and Gao, Weibo and Liu, Qi and Li, Jiatong and Zhao, Guanhao and Zhang, Zheng and Huang, Zhenya and Zhu, Mengxiao and Wang, Shijin and Tong, Wei and others},
  journal={arXiv preprint arXiv:2407.05458},
  year={2024}
}

@inproceedings{huang2019exploring,
  title={Exploring multi-objective exercise recommendations in online education systems},
  author={Huang, Zhenya and Liu, Qi and Zhai, Chengxiang and Yin, Yu and Chen, Enhong and Gao, Weibo and Hu, Guoping},
  booktitle={Proceedings of the 28th ACM international conference on information and knowledge management},
  pages={1261--1270},
  year={2019}
}

@inproceedings{10.1145/3357384.3357995,
author = {Huang, Zhenya and Liu, Qi and Zhai, Chengxiang and Yin, Yu and Chen, Enhong and Gao, Weibo and Hu, Guoping},
title = {Exploring Multi-Objective Exercise Recommendations in Online Education Systems},
year = {2019},
booktitle = {CIKM '19},
isbn = {9781450369763},
publisher = {Association for Computing Machinery},
address = {New York, NY, USA},
url = {https://doi.org/10.1145/3357384.3357995},
doi = {10.1145/3357384.3357995},
pages = {1261–1270},
numpages = {10},
location = {Beijing, China},
series = {CIKM '19}
}

@inproceedings{bi2020quality,
  title={Quality meets diversity: A model-agnostic framework for computerized adaptive testing},
  author={Bi, Haoyang and Ma, Haiping and Huang, Zhenya and Yin, Yu and Liu, Qi and Chen, Enhong and Su, Yu and Wang, Shijin},
  booktitle={2020 IEEE International Conference on Data Mining (ICDM)},
  pages={42--51},
  year={2020},
  organization={IEEE}
}

@inproceedings{DISCO,
title={DISCO: A Hierarchical Disentangled Cognitive Diagnosis Framework for Interpretable Job Recommendation},
author={Yu, Xiaoshan and Qin, Chuan and Zhang, Qi and Zhu, Chen and Ma, Haiping and Zhang, Xingyi and Zhu, Hengshu},
booktitle={2024 IEEE International Conference on Data Mining (ICDM)},
  year={2024},
  organization={IEEE}
}

@article{brown2020language,
  title={Language models are few-shot learners},
  author={Brown, Tom and Mann, Benjamin and Ryder, Nick and Subbiah, Melanie and Kaplan, Jared D and Dhariwal, Prafulla and Neelakantan, Arvind and Shyam, Pranav and Sastry, Girish and Askell, Amanda and others},
  journal={Advances in neural information processing systems},
  volume={33},
  pages={1877--1901},
  year={2020}
}

@article{kojima2022large,
  title={Large language models are zero-shot reasoners},
  author={Kojima, Takeshi and Gu, Shixiang Shane and Reid, Machel and Matsuo, Yutaka and Iwasawa, Yusuke},
  journal={Advances in neural information processing systems},
  volume={35},
  pages={22199--22213},
  year={2022}
}

@inproceedings{ma2025debate,
  title={Debate on graph: a flexible and reliable reasoning framework for large language models},
  author={Ma, Jie and Gao, Zhitao and Chai, Qi and Sun, Wangchun and Wang, Pinghui and Pei, Hongbin and Tao, Jing and Song, Lingyun and Liu, Jun and Zhang, Chen and others},
  booktitle={Proceedings of the AAAI Conference on Artificial Intelligence},
  volume={39},
  pages={24768--24776},
  year={2025}
}

@article{wang2020instructions,
  title={Instructions and guide for diagnostic questions: The neurips 2020 education challenge},
  author={Wang, Zichao and Lamb, Angus and Saveliev, Evgeny and Cameron, Pashmina and Zaykov, Yordan and Hern{\'a}ndez-Lobato, Jos{\'e} Miguel and Turner, Richard E and Baraniuk, Richard G and Barton, Craig and Jones, Simon Peyton and others},
  journal={arXiv preprint arXiv:2007.12061},
  year={2020}
}

@article{liu2023xes3g5m,
  title={Xes3g5m: A knowledge tracing benchmark dataset with auxiliary information},
  author={Liu, Zitao and Liu, Qiongqiong and Guo, Teng and Chen, Jiahao and Huang, Shuyan and Zhao, Xiangyu and Tang, Jiliang and Luo, Weiqi and Weng, Jian},
  journal={Advances in Neural Information Processing Systems},
  volume={36},
  pages={32958--32970},
  year={2023}
}

@inproceedings{long2022improving,
  title={Improving knowledge tracing with collaborative information},
  author={Long, Ting and Qin, Jiarui and Shen, Jian and Zhang, Weinan and Xia, Wei and Tang, Ruiming and He, Xiuqiang and Yu, Yong},
  booktitle={Proceedings of the fifteenth ACM international conference on web search and data mining},
  pages={599--607},
  year={2022}
}

@inproceedings{rasley2020deepspeed,
  title={Deepspeed: System optimizations enable training deep learning models with over 100 billion parameters},
  author={Rasley, Jeff and Rajbhandari, Samyam and Ruwase, Olatunji and He, Yuxiong},
  booktitle={Proceedings of the 26th ACM SIGKDD international conference on knowledge discovery \& data mining},
  pages={3505--3506},
  year={2020}
}

@article{liu2019roberta,
  title={Roberta: A robustly optimized bert pretraining approach},
  author={Liu, Yinhan and Ott, Myle and Goyal, Naman and Du, Jingfei and Joshi, Mandar and Chen, Danqi and Levy, Omer and Lewis, Mike and Zettlemoyer, Luke and Stoyanov, Veselin},
  journal={arXiv preprint arXiv:1907.11692},
  year={2019}
}

@inproceedings{xiao2024c,
  title={C-pack: Packed resources for general chinese embeddings},
  author={Xiao, Shitao and Liu, Zheng and Zhang, Peitian and Muennighoff, Niklas and Lian, Defu and Nie, Jian-Yun},
  booktitle={Proceedings of the 47th international ACM SIGIR conference on research and development in information retrieval},
  pages={641--649},
  year={2024}
}

@article{hu2022lora,
  title={Lora: Low-rank adaptation of large language models.},
  author={Hu, Edward J and Shen, Yelong and Wallis, Phillip and Allen-Zhu, Zeyuan and Li, Yuanzhi and Wang, Shean and Wang, Lu and Chen, Weizhu and others},
  journal={ICLR},
  volume={1},
  number={2},
  pages={3},
  year={2022}
}

@article{gan2020modeling,
  title={Modeling learner’s dynamic knowledge construction procedure and cognitive item difficulty for knowledge tracing},
  author={Gan, Wenbin and Sun, Yuan and Peng, Xian and Sun, Yi},
  journal={Applied Intelligence},
  volume={50},
  number={11},
  pages={3894--3912},
  year={2020},
  publisher={Springer}
}

@article{choi2020predicting,
  title={Predicting the difficulty of EFL tests based on corpus linguistic features and expert judgment},
  author={Choi, Inn-Chull and Moon, Youngsun},
  journal={Language Assessment Quarterly},
  volume={17},
  number={1},
  pages={18--42},
  year={2020},
  publisher={Taylor \& Francis}
}

@article{fan1998item,
  title={Item response theory and classical test theory: An empirical comparison of their item/person statistics},
  author={Fan, Xitao},
  journal={Educational and psychological measurement},
  volume={58},
  number={3},
  pages={357--381},
  year={1998},
  publisher={Sage Publications Sage CA: Thousand Oaks, CA}
}

@article{qwen3embedding,
  title={Qwen3 Embedding: Advancing Text Embedding and Reranking Through Foundation Models},
  author={Zhang, Yanzhao and Li, Mingxin and Long, Dingkun and Zhang, Xin and Lin, Huan and Yang, Baosong and Xie, Pengjun and Yang, An and Liu, Dayiheng and Lin, Junyang and Huang, Fei and Zhou, Jingren},
  journal={arXiv preprint arXiv:2506.05176},
  year={2025}
}

\appendix

\section{Appendix }

\subsection{Details about the Datasets}\label{subsec:Appendix A.1}
In this research, we selected two real-world online education datasets, XES3G5M and NIPS34, which vary in size and sparsity, to comprehensively evaluate the performance of our proposed method across different practical application scenarios. The statistical information of the datasets is presented in Table \ref{dataset_info}.

\textbf{NIPS34}: A sub-dataset of the NeurIPS Education Challenge, containing Tasks 3 and 4, is a powerful resource tailored to the advancement of educational data analytics and machine learning applications within the education field. The dataset comprises crowdsourced diagnostic mathematics exercises collected from the Eedi educational platform between September 2018 and May 2020, targeting students from elementary through high school. Task 3 aims to accurately predict which exercises are of high quality, while Task 4 seeks to determine a personalized sequence of exercises for each student that optimally predicts their responses. NIPS34  contains a variety of features including interaction logs, the content of the exercises in picture format, the names of the KCs and their structured annotations, which allow for a comprehensive analysis of learning behaviors and outcomes. NIPS34 contains more than 1,300,000 records, providing a wealth of high-quality data suitable for multiple types of tasks, making it an invaluable tool for conducting cognitive diagnosis research.

\textbf{XES3G5M}: XES3G5M is a large-scale dataset comprising over five million interactions collected from more than 18,000 third-grade students responding to approximately 8,000 math exercises. It includes extensive auxiliary information related to the exercises and their associated KCs. Sourced from a real-world online mathematics learning platform, XES3G5M not only encompasses the largest number of KCs within the mathematics domain but also provides the most comprehensive contextual information. This includes hierarchical KC relationships, exercise types, Chinese textual content and analyses, as well as timestamps of student responses.
\begin{table}[!ht]
    \centering
    \setlength{\tabcolsep}{8pt}    
    \begin{tabular}{c|c c}         
        \hline  
         & XES3G5M & NIPS34 \\
         \hline
        \#Student & 11,453 & 4,918 \\
        \#Exercise & 7,652 & 948 \\
        \#KC & 1,175 & 86 \\
        \#Log & 5,139,044 & 1,382,727 \\
        \#Log per student & 448.7 & 281.2 \\
        \#Log per exercise & 671.6 & 1,458.6 \\
        \#Sparsity(\%) & 5.1 & 29.7 \\
        \hline
    \end{tabular}
    \caption{Statistics of datasets.}
    \label{dataset_info}
\end{table}
In this experiment, we use five-fold cross validation to ensure the reliability of the experimental results. We use the full data set of NIPS34. As for the XES dataset, due to the fact that there are too many record exercises, has done the following treatment, and only 2000 students' corresponding records of doing the exercises are kept in each fold. 

\subsection{Knowledge Structure of the Datasets}\label{subsec:Appendix A.2}
\begin{figure*}[t!]
\centering
\includegraphics[width=\textwidth]{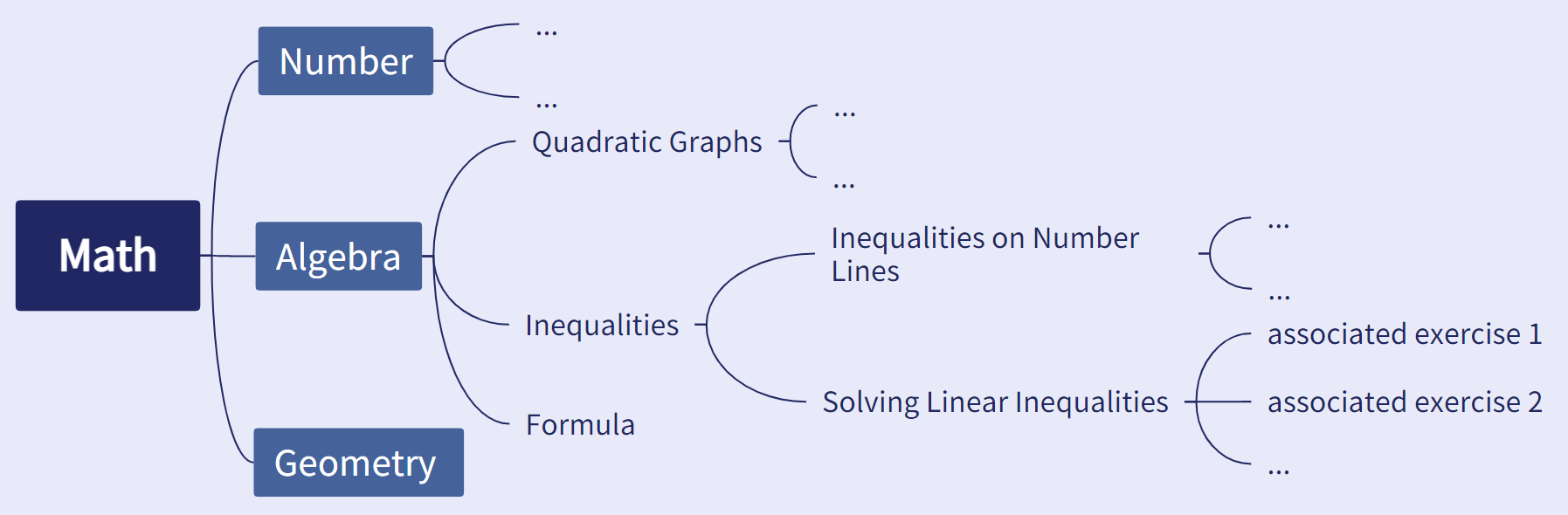} 
\caption{The schematic diagram of the hierarchical structure of knowledge concepts in the NIPS34 dataset.}
\label{kc_struct}
\vspace{-1.0em}
\end{figure*}

In practical educational scenarios, KCs are typically organized into
 hierarchical tree structures, as illustrated in Figure \ref{kc_struct}. In a top-down manner, the root node $\mathcal{K}^0 = \left\langle \mathbb{K}, \mathcal{R} \right\rangle = \bigcup_{i=1}^{M^1} \mathcal{D}_i$ represents full dataset or its knowledge system (e.g., ``Math" in NIPS34), consisting of the complete KC set $\mathbb{K}$ and the branch set $\mathcal{R}$. Assuming there are a total of $M_0^1$ nodes at depth 1, each of them is considered a distinct domain $\mathcal{D}_i = \left\langle \mathbb{K}^1_i, \mathcal{R}^1_i \right\rangle$, $\forall i \in [1,M_0^1]$, where $\mathbb{K}^1_i = \{k_{ij}^l \,|\, j \in [1,2,...,M_{i}^l],\, l \in [1,2,...,max\_depth] \}$, and $\mathcal{R}^1_i = \{(k_{ip},\, k_{iq}) \,|\,k_{ip},\,k_{iq} \in \mathbb{K}^1_i\}$, $(k_{ip},\, k_{iq})$ is a directed edge of the tree. Note that knowledge across different domains is entirely isolated, that is, $\mathbb{K}^1_i \cap \mathbb{K}^1_j = \emptyset$, $\forall i \neq j$. The hierarchical relationship(s) from the root to the leaf node(s) representing fine-grained terminal KCs associated with exercise $v$, is called its ``knowledge route" $\mathcal{K}_v$.


\subsection{The definition of the Cold Start problem}\label{subsec:Appendix A.3}
\textbf{Zero-Shot setting.} Consistent with previous studies on cognitive diagnosis, Zero-Shot means that the dataset is divided into two isolated parts in a certain dimension for model training and testing, respectively. Here, our proposed LMCD focuses primarily on exercise and domain-level tasks.

\textbf{Exercise cold-start.} As a classic challenge in CDMs, exercise cold-start refers to the situation in which a new exercise is first introduced into the exercise pool, and due to the absence of historical student response data, it becomes difficult to accurately estimate the exercise's diagnostic properties. Following TechCD \cite{gao2023leveraging}, we randomly divide dataset into warm ($\mathcal{H}$) and cold ($\mathcal{C}$) subsets on exercise-level. In $\mathcal{H}$, there are students $\mathcal{U_{H}} = \{ u_1,u_2, \cdots, u_{|\mathcal{U_{H}}|} \}$, exercises $\mathcal{V_{H}} = \{ v_1, v_2, \cdots, v_{|\mathcal{V_{H}}|} \}$, KCs $\mathcal{K_{H}} = \{ k_1,k_2, \cdots, k_{|\mathcal{K_{H}}|} \}$, and response logs $\mathcal{R}_\mathcal{H} = \left\{(u,v,\mathcal{K}_{v},y_{uv})\mid u \in \mathcal{U_{H}}, v \in \mathcal{V_{H}}\right\}$ where $y_{uv} \in \{0,1\}$ indicates the correctness of student $u$ on exercise $v$. Similarly, we define $\mathcal{U_{C}}$, $\mathcal{V_{C}}$, $\mathcal{K_{C}}$, and $\mathcal{R}_\mathcal{C} = \left\{(u,v,\mathcal{K}_{v},y_{uv})\mid u \in \mathcal{U_{C}}, v \in \mathcal{V_{C}}\right\}$ for $\mathcal{C}$. Note that the KCs overlap ($\mathcal{K_{H}} \cap \mathcal{K_{C}} \neq \emptyset$) while the exercise sets are completely disjoint ($\mathcal{V_{H}} \cap \mathcal{V_{C}} = \emptyset$). In other words, $\mathcal{H}$ and $\mathcal{C}$ share knowledge indirectly through domain-level.

\textbf{Domain cold-start.} Equivalent to the domain-level zero-shot cognitive diagnosis (DZCD) from ZeroCD \cite{gao2024zero}, in a more constrained way than exercise cold-start, we divide $\mathcal{K}^0$ at depth 1 into warm domain(s) $\mathcal{H}$ and cold domain(s) $\mathcal{C}$. The corresponding KC sets $\mathcal{K_{\mathcal{H}}} = \bigcup_{i=1}^{M_\mathcal{H}} \mathcal{K}_i^1$ and $\mathcal{K_{\mathcal{C}}} = \bigcup_{i=M_\mathcal{H}}^{M} \mathcal{K}_i^1$ satisfy $\mathcal{K_{\mathcal{H}}} \cap \mathcal{K_{\mathcal{C}}} = \emptyset$, $\mathcal{V}_{\mathcal{H}} \cap \mathcal{V}_{\mathcal{C}} = \emptyset$, and $\mathcal{U_{C}} \subseteq \mathcal{U_{H}}$, meaning that $\mathcal{H}$ and $\mathcal{C}$ are isolated at both exercise and domain levels.

\subsection{Related Work}\label{subsec:Appendix related work}



{\bf Cognitive Diagnosis.} Cognitive diagnosis in education has evolved from traditional psychometric approaches to deep learning models. Early methods like IRT \cite{lord1952theory} and MIRT \cite{reckase200618} model student-exercise interactions through logistic functions, while DINA \cite{de2009dina} introduce slip and guess parameters. These foundational approaches offered interpretability but limited expressiveness. NCDM \cite{wang2020neural} mark a significant advancement by leveraging neural networks to capture more complex knowledge representations. Recent research has expanded the modeling scope by incorporating additional factors such as emotional states \cite{wang2024boosting}, unlabeled data \cite{chen2023disentangling}, and response time \cite{ma2025ad4cd}. Despite advances, most methods struggle in cold-start scenarios with limited data.

{\bf Cold-start in Cognitive Diagnosis.} Solutions primarily fall into two categories, graph-based and NLP-based approaches. Graph-based methods \cite{liu2024inductive,gao2023leveraging,gao2024zero} leverage structural relationships between educational entities. TechCD \cite{gao2023leveraging} leverage tailored knowledge concept graphs linking different domains but requires overlapping students. ZeroCD \cite{gao2024zero} utilizes early bird students in target domains to learn transferable cognitive signals, though this requirement limits practical application. NLP-based approaches show promising results by utilizing language models like BERT to encode exercise text \cite{gao2023leveraging,gao2024zero}, but face limits such as oversimplifying exercise difficulty and misreading unclear texts, leading to inaccuracies in cross-domain settings.

{\bf LLMs in Cognitive Diagnosis}. LLMs have revolutionized natural language processing with advanced comprehension  \cite{brown2020language} and reasoning capabilities \cite{kojima2022large,ma2025debate}, but their integration with cognitive diagnosis remains nascent. Current approaches include LRCD \cite{Liu2025LRCD}, which embeds students, exercises, and concepts into a unified language space; and KCD \cite{dong2025knowledge}, which exploits LLMs' reasoning ability to generate diagnostic information for students and exercises. However, these approaches' simplistic use of LLMs limits their effectiveness in cold-start scenarios.


\subsection{Details about Baselines}\label{subsec:Appendix A.4}
In this subsection, we provide a detailed explanation of the specific modifications made to
 each baseline experiment based on its original implementation to adapt it to the data split used in the specific cold-start scenario of this research:

\textbf{IRT/MIRT}: We adopt a three-parameter logistic (3PL) form for the interaction
 function. The temperature coefficient is maintained at its original setting of 1.703, and the feature dimension of MIRT is set to 4. Meanwhile, considering the specific types of the exercises, we set the upper bound of the guess coefficient to 0.5. In addition, Xavier initialization is applied to all embedding layers in the models.  

\textbf{NeuralCD}: We employ the default settings without any modifications to ensure the
 consistency of the experiment results.

\textbf{TechCD}: The model architecture remains consistent with the original setup, with the
 addition of optional IRT and MIRT predict heads. Furthermore, undirected edges characterizing the similarity between fine-grained ($depth \geq 3$) KCs under the same parent node are also introduced in the construction of the Knowledge Concept Graph (KCG) for the NIPS34 and XES3G5M datasets, based on the provided KC tree structure.

\textbf{NLP-base}: We employ language models with frozen weights to embed the textual information of students and exercises, including descriptions generated by LLMs, thereby replacing the ID embedding layers in IRT/MIRT and NCDM. Experiments are conducted using BERT, RoBERTa, BGE and Qwen3-Embedding as text embedders. Specifically, we use bert-base-uncased/chinese, bge-large-en/zh-v1.5, xlm-roberta-base and Qwen/Qwen3-Embedding-4B for adaptation to datasets in different languages.

\textbf{KCD}: We use Qwen-Plus for information extraction and diagnosis and modify the prompts as follows. First, input interactions are truncated up to 20 to avoid excessively long prompts, while contents related to interactions are removed from the prompts that generate descriptions for cold-start data. Then, the text embedder employed is bge-large-en/zh-v1.5, and a Chinese version of the prompt was added for the XES3G5M dataset.

\subsection{Details about Knowledge Diffusion}\label{subsec:Appendix A.5}
In order to generate descriptions for each KC, we utilize the exact format of prompt as follows:

\begin{tcolorbox}[
  colback=blue!5!white,    
  colframe=blue!75!black,  
  title={Prompt for KC Description Generation}, 
  fonttitle=\color{white}
]

{\small
System Prompt: ``If you are a seasoned math teacher, you need to generate explanations for each knowledge concept in a knowledge graph. I will provide you with the name of the knowledge concept and corresponding example problems, as well as names and example problems of distractor knowledge concepts that are not equivalent to it. Please carefully compare them and generate a core explanation for each knowledge concept."

------------------------------------------------

KC Name: $k_{target}$

Example exercises for $k_{target}$:

$Q_{target}$

------------------------------------------------

Distractor KCs

KC Name: $k^1_{neg}$

Example exercises for $k^1_{neg}$:

$Q^{1}_{neg}$

...

KC Name: $k^n_{neg}$

Example exercises for $k^N_{neg}$: 

$Q^{N}_{neg}$
}
\end{tcolorbox}


\textbf{Enrichment and robustness.} As shown in Case 1, when describing the KC ``Fractions, Decimals and Percentage Equivalence", the inclusion of distractors effectively prevents the LLM from mechanically enumerating definitions based solely on the target KC's name. Furthermore, it avoids an excessive focus on the content of example exercises leading to redundant information. In summary, the inclusion of distractors not only enriches the prompt but also enhances the robustness of the generated results by mitigating sensitivity to the selection of specific example exercises.

\textbf{Abstraction and discrimination.} Besides, as shown in Case 2, which describe the KC ``Angles", the incorporation of distractors results in more concise descriptions which predominantly emphasize the knowledge and skill elements rather than specific examples. In other words, distractor KCs facilitate the abstraction of concrete exercises, thereby accentuating the differences between similar KCs at knowledge and skill level. This abstraction contributes to higher precision and discrimination of the generated descriptions.


\begin{tcolorbox}[
  colback=blue!5!white,    
  colframe=blue!75!black,  
  title={KC Descriptions: Case 1}, 
  fonttitle=\color{white}
]

{\small
Target KC: \textbf{Fractions, Decimals and Percentage Equivalence}

------------------------------------------------

\textbf{With distractor KCs:}

``Fractions involve understanding parts of a whole and performing mathematical operations such as addition, subtraction, multiplication, and division using fractional numbers. This includes interpreting fractions visually and solving problems that require manipulation of fractions in equations or patterns."
        
------------------------------------------------

\textbf{Without distractor KCs:}

``Fractions represent parts of a whole and are written as a numerator over a denominator. The numerator indicates how many parts are being considered, while the denominator shows the total number of equal parts the whole is divided into. Operations with fractions follow specific rules:

1. **Multiplication**: Multiply the numerators and denominators separately, then simplify if necessary.

    - Example: $\frac{1}{3} \times (-9) = \frac{-9}{3} = -3$.

2. **Division**: ..."
}

\end{tcolorbox}


\begin{tcolorbox}[
  colback=blue!5!white,    
  colframe=blue!75!black,  
  title={KC Descriptions: Case 2}, 
  fonttitle=\color{white}
]

{\small
Target KC: \textbf{Angles}

------------------------------------------------

\textbf{With distractor KCs:}

``Angles are geometric measures that represent the amount of turn between two intersecting lines or line segments. They can be measured in degrees, estimated visually, compared in size, or determined using tools like protractors. Understanding angles involves recognizing different types (acute, obtuse, right, etc.) and performing operations such as addition, subtraction, or fraction-based calculations involving turns."

------------------------------------------------

\textbf{Without distractor KCs:}

``An \textbf{Angle} is a measure of rotation or the amount of turn between two intersecting lines or rays. Angles are typically measured in degrees ($^{\circ}$), with a full turn equaling $360^{\circ}$. A half-turn corresponds to $180^{\circ}$, and a quarter-turn equals $90^{\circ}$, which is also called a right angle. Angles can be classified based on their size:

\textbf{Acute angles} are less than $90^{\circ}$.

\textbf{Right angles} are ...

To estimate or measure angles, tools such as protractors are used. Estimation involves comparing the given angle to known reference angles (e.g., $90^{\circ}$, $180^{\circ}$). In cases where direct measurement is not possible, logical reasoning or comparison may help determine relationships between angles."
}

\end{tcolorbox}

\subsection{Detailed Experimental Settings and Hyperparameters}\label{sec:appendix_hyperparameters}
To ensure reproducibility, we detail our experimental setup and hyperparameters. Specifically, we utilized Qwen-Plus to generate descriptions for Knowledge Concepts (KCs) and adopted Qwen2.5-1.5B-base \cite{yang2024qwen2} as our foundation model. Training was optimized using the DeepSpeed \cite{rasley2020deepspeed} framework, employing Low-Rank Adaptation (LoRA) with a rank of 16 for parameter-efficient fine-tuning. The model was trained for up to 10 epochs with a global batch size of 64. To stabilize training and prevent overfitting, we set the initial learning rate to 0.0001, decayed via a linear scheduler, and applied a dropout rate of 0.1.


\subsection{Detailed Results for Domain Cold-Start (RQ2)}\label{subsec:Appendix_add_exp}
In this appendix, we provide the detailed experimental settings and results for the domain cold-start scenarios (RQ2) discussed in the main text. 

We conducted these challenging cross-domain experiments on the NIPS34 dataset. It is worth noting that due to the sparsity of knowledge embeddings in NCDM, which cannot be effectively transferred to cross-domain scenarios, our evaluation here focuses exclusively on the IRT and MIRT models. The comprehensive results are reported in Table \ref{kccold_exp}.

From the results, we draw several key observations:
\begin{itemize}
    \item \textbf{Performance Comparison:} In domain cold-start scenarios, our proposed LMCD achieved nearly optimal results in both the Algebra and Geometry domains. Meanwhile, KCD performed best in the Number domain. The effectiveness of KCD stems from its use of Large Language Models (LLMs) to generate textual summaries of both students' problem-solving behaviors and exercises' user-interaction statistics, which are then transferred to Cognitive Diagnosis Models (CDMs) during training. Nevertheless, LMCD remains highly competitive by leveraging the textual descriptions of KCs while effectively incorporating students' cognitive information into the problem representations.
    \item \textbf{Limitations of Graph-based Methods:} By contrast, the graph-based TechCD method failed completely in these cross-domain scenarios. This failure is primarily because the cold-start data is isolated from the training set at both the KC and exercise levels. Consequently, no transferable knowledge between domains can be captured through the graph structure.
\end{itemize}

In summary, the detailed experimental results further validate that LMCD demonstrates superior robustness and adaptability in domain cold-start scenarios compared to existing baselines.

\begin{table*}[!ht]
\centering
\setlength{\tabcolsep}{4pt}  
\begin{tabular}{l l ccc ccc ccc}
\toprule
\multirow{2}{*}{CDM} & \multirow{2}{*}{Method} & \multicolumn{3}{c}{Number as Target} & \multicolumn{3}{c}{Algerbra as Target} & \multicolumn{3}{c}{Geometry as Target} \\
\cmidrule(lr){3-5} \cmidrule(lr){6-8} \cmidrule(lr){9-11}
& & ACC$\uparrow$ & AUC$\uparrow$ & RMSE$\downarrow$ & ACC$\uparrow$ & AUC$\uparrow$ & RMSE$\downarrow$ & ACC$\uparrow$ & AUC$\uparrow$ & RMSE$\downarrow$ \\
\midrule
\multirow{8}{*}{IRT} 
& Oracle &0.7215	&0.7931	&0.4278	&0.7108	&0.7691	&0.4379	&0.7172	&0.7851	&0.4329 \\
& Random &0.4932	&0.5022	&0.5002	&0.5019	&0.5015	&0.5001	&0.4864	&0.4929	&0.5004 \\
& Bert  &0.6092 &0.6573 &0.4903 &0.6251 &0.6689 &0.4800 &0.6120 &0.6462 &0.4863 \\
& Roberta  &\textbf{0.6369} &0.6824 &\textbf{0.4720} &0.6483 &0.6982 &\underline{0.4692} &\underline{0.6224} &0.6718 &\underline{0.4816} \\
& Bge  &0.6204 &0.6521 &0.4957 &0.5671 &0.6496 &0.5531 &0.5979 &0.6498 &0.5145 \\
& Qwen3E &0.6230 &0.6766 &0.4860 &0.6224 &0.6848 &0.4846 &0.6174 &0.6638	&0.4965 \\
& TechCD &0.4827 &0.5151 &0.5059 &0.4936 &0.5071 &0.5125 &0.4959 &0.5046	&0.5084 \\
& KCD  &\underline{0.6357} &\textbf{0.6945} &0.4880 &\underline{0.6505} &\textbf{0.7091} &0.4857 &0.6271 &\underline{0.6726}	&0.4888 \\
& LMCD  &0.6336 &\underline{0.6837} &\underline{0.4726} &\textbf{0.6518} &\underline{0.7049}	&\textbf{0.4658} &\textbf{0.6363} &\textbf{0.6843} &\textbf{0.4758} \\
\midrule
\multirow{8}{*}{MIRT} 
& Oracle &0.7223	&0.7942	&0.4272	&0.7116	&0.7694	&0.4377	&0.7180	&0.7860	&0.4324 \\
& Random &0.4354	&0.4938	&0.5328	&0.5064	&0.4951	&0.5154 &0.4770	&0.5010	&0.5224 \\
& Bert  &0.6339 &0.6712 &0.4814 &0.6318 &0.6818 &0.4761 &0.6212 &0.6683 &\underline{0.4828} \\
& Roberta  &0.6195 &0.6432 &0.5125 &\underline{0.6524} &0.7045 &\underline{0.4659} &0.6096 &0.6339 &0.5308 \\
& Bge  &0.6341 &0.6681 &0.4842 &0.6366 &0.6904 &0.4819 &0.6115 &0.6682 &0.4879 \\
&Qwen3E &\underline{0.6402} &0.6808 &0.4826 &0.6420 &0.6884 &0.4717 &\underline{0.6214} &0.6658	&0.4880 \\
& TechCD &0.5646 &0.5267 &0.4956 &0.4936 &0.5153 &0.5009 &0.5231 &0.5014	&0.4995 \\
& KCD  &\textbf{0.6500}	&\textbf{0.6974} &\textbf{0.4715} &0.5922	&\textbf{0.7120}	&0.4992	&0.6025	&\underline{0.6754} &0.4994 \\
& LMCD  &0.6269	&\underline{0.6888}  &\underline{0.4778} &\textbf{0.6551} &\underline{0.7082} &\textbf{0.4656} &\textbf{0.6277} &\textbf{0.6821} &\textbf{0.4783} \\
\bottomrule
\end{tabular}
\caption{Performance Comparison of Domain Cold Start with IRT and MIRT on NIPS34. The best performance is highlighted in bold, and the the second-best performances is underlined.}
\label{kccold_exp}
\end{table*}

\subsection{Additional Ablation Studies}\label{sec:appendix_ablation}
{\bf Impact of model architecture.} 
Within the LMCD framework, we analyzed the performance of different model architectures and sizes. As shown in Table \ref{abs:diffmodel}, we conducted experiments on decoder-based Qwen2.5 models across different sizes and adapted encoder-based RoBERTa models, finding that the decoder-based LMCD framework achieved superior performance compared to the encoder-based variant. This highlights the advantage of causal attention in effectively aggregating semantic information over specific ranges (i.e., exercise-level or exercise-and-student-level), which contributes to the enhanced performance of the LMCD model. Additionally, we assessed models of varying sizes (0.5B to 7B), finding consistent performance improvements from 0.5B to 3B, followed by a slight decline at 7B. This suggests that while larger models offer richer semantic information, they may dilute cognitive features, leading to diminished performance at excessive scales.

\begin{table}[ht]
\centering
\begin{tabular}{lccc}
\hline
\textbf{Model(LMCD)} & \textbf{ACC} & \textbf{AUC} & \textbf{RMSE} \\ \hline
Roberta-0.3b &0.6725 &0.7293 &0.4579 \\
Qwen2.5-0.5b & 0.6808 & 0.7414 & 0.4538 \\
Qwen2.5-1.5b & 0.6813	 & 0.7440 & 0.4525 \\
Qwen2.5-3b & 0.6842 & 0.7434 & 0.4513 \\
Qwen2.5-7b   &0.6783 &0.7377 &0.4538 \\ \hline
\end{tabular}
\caption{Experimental Results across Different LM Architectures and Model Sizes}
\label{abs:diffmodel}
\end{table}

{\bf Impact of knowledge encoding strategies.} Figure \ref{abs_text} demonstrates the effectiveness of our ``knowledge diffusion'' approach across three experimental conditions based on different inputs to Eq.\ref{emb_kv}: {\bf Q} (exercise text only), {\bf KQ} (exercise text with knowledge concept labels), and {\bf DKQ} (adding LLM-generated KC descriptions to KQ). Results show that enriching encoding content generally improves performance across models, with LMCD and Roberta showing consistent gains in AUC. BERT, however, performs worse with KQ than with Q alone, likely due to the ambiguity of KC labels without context. This pattern validates the necessity of our ``knowledge diffusion'' approach, which provides detailed KC descriptions rather than relying solely on potentially ambiguous concept labels.

\begin{figure}[!ht]
\centering
\includegraphics[width=0.45\textwidth,height=0.15\textheight]{yzh/img/ABS_Gen4_up2.png} 
\caption{Impact of knowledge encoding strategies.}
\label{abs_text} 
\end{figure}

\end{document}